\documentclass[letterpaper, 10 pt, conference]{ieeeconf}  

\IEEEoverridecommandlockouts                              

\makeatletter
\let\NAT@parse\undefined
\makeatother

\usepackage{times}
\usepackage[hidelinks,bookmarks=true,backref=section]{hyperref}
\usepackage{times}
\usepackage{graphicx}
\usepackage{xcolor}
\usepackage{algorithm}
\usepackage{algorithmicx}
\usepackage[noend]{algpseudocode}
\usepackage{amsmath}
\usepackage{booktabs}
\usepackage{graphicx}
\usepackage{amssymb} 
\usepackage{multicol}
\usepackage{bbm}
\usepackage{wrapfig}
\usepackage{arydshln}
\usepackage[font=small,labelfont=bf]{caption}
\usepackage{balance}
\usepackage{etoolbox}
\usepackage{subcaption}
\newcommand{\rebuttal}[1]{\textcolor{black}{#1}}
\newcommand{\citet}[1]{\cite{#1}}
\newcommand{\citep}[1]{\cite{#1}}

\title{\LARGE \bf
RecoveryChaining: \\ Learning Local Recovery Policies for Robust Manipulation
}

\author{Shivam Vats$^{1}$, Devesh K. Jha$^{2}$, Maxim Likhachev$^{3}$, Oliver Kroemer$^{3}$ and Diego Romeres$^{2}$
\thanks{$^{\dagger}$ The Robotics Institute, CMU
        {\tt\small \{svats, okroemer, mlikhach\}@andrew.cmu.edu}}%
\thanks{$^{\ddagger}$ MERL, Cambridge, MA 02139
        {\tt\small \{jha,romeres\}@merl.com}}%
}
\begin{document}
\thispagestyle{empty}
\pagestyle{empty}

\twocolumn[{%
    \renewcommand\twocolumn[1][]{#1}%
    \maketitle
    \begin{center}
        \centering
        \includegraphics[width=1\textwidth]{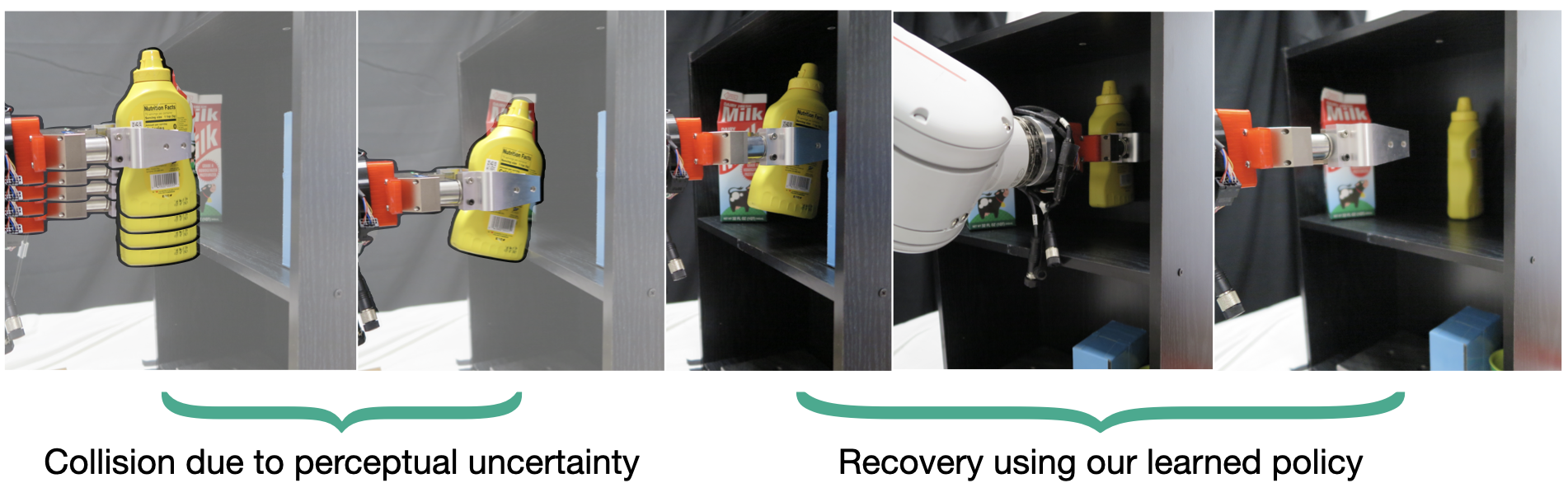}
        \captionof{figure}{
            Due to uncertainty in the grasp of the object, the robot ends up making contact with the shelf during execution resulting in a collision as well as an in-hand slip failure. However, using the proposed recovery chaining framework, the robot recovers from the collision state and then hands over control to the nominal place skill. The learnt recovery also allows the robot to correct the slip by intentionally making contact with the shelf wall. Note that the policy is trained entirely in simulation. [Best viewed in color].
        }
        \label{fig:robot_execution}
    \end{center}%
    }]
\footnotetext[1]{Brown University,
        {\tt\small shivam\_vats@brown.edu}}%
\footnotetext[2]{Mitsubishi Electric Research Labs,
        {\tt\small \{jha,romeres\}@merl.com}%
}%
\footnotetext[3]{Robotics Institute, Carnegie Mellon University,
{\tt\small \{maxim, okroemer\}@cs.cmu.edu}}%

\begin{abstract}
Model-based planners and controllers are commonly used to solve complex manipulation problems as they can efficiently optimize diverse objectives and generalize to long horizon tasks. However, they often fail during deployment due to noisy actuation, partial observability and imperfect models. To enable a robot to recover from such failures, we propose to use hierarchical reinforcement learning to learn a recovery policy. The recovery policy is triggered when a failure is detected based on sensory observations and seeks to take the robot to a state from which it can complete the task using the nominal model-based controllers. Our approach, called RecoveryChaining, uses a hybrid action space, where the model-based controllers are provided as additional \emph{nominal} options  which allows the recovery policy to decide how to recover, when to switch to a nominal controller and which controller to switch to even with \emph{sparse rewards}. We evaluate our approach in three multi-step manipulation tasks with sparse rewards, where it learns significantly more robust recovery policies than those learned by baselines. We successfully transfer recovery policies learned in simulation to a physical robot to demonstrate the feasibility of sim-to-real transfer with our method.
\end{abstract}
\section{Introduction}
A robot trying to tidy up a house is confronted with a myriad of possible failures.
It might pick up a half-read \emph{`Planning Algorithms'} book from the table and try to place it in the top shelf but fail to see the objects already inside.
How does it respond when the book bumps into the clutter and starts slipping out of its hand?
One potential recovery behavior would be to first fix the slip by pushing the book against the shelf and then repositioning it for another attempt.
Humans can quickly come up with robust strategies to deal with such failures, often with just partial information.
For example, we often use our sense of touch to extract objects from a bag when we don't have a clear view of its contents, and we use our quick reflexes to recover from slipping on a patch of ice.

However, popular model-based planning approaches~\cite{mason2001mechanics,kroemer2021review} struggle to generate such behaviors on-the-fly because of their reliance on an approximate model of the environment.
This reliance leads to failures when the robot is faced with large inaccuracies in the model during deployment~\cite{lagrassa2020learning}.
Hence, robots are usually deployed with recovery behaviors to gracefully handle potential failures.
Common recovery strategies include retrying the previous step~\cite{ebert2018robustness}, backtracking~\cite{wang2019robust} and hand-designed corrective actions~\cite{sundaresan2021untangling}.
These heuristic strategies can be sub-optimal and  require significant manual engineering effort.
In this paper, we propose to use reinforcement learning (RL)~\cite{sutton2018reinforcement,lillicrap2015continuous} to automate the discovery of robust recovery behaviours for multi-step manipulation.
RL is capable of discovering and learning complex robot skills~\cite{haarnoja2023learning,zhou2023learning} but is limited by the twin problems of (1) high sample complexity and (2) significant reward shaping.

To address this, \rebuttal{we propose a novel hierarchical reinforcement learning (HRL) formulation \textbf{RecoveryChaining} for recovery learning that is much more sample efficient than flat RL and can solve challenging manipulation problems even with a \emph{sparse reward}.}
Our main idea is to use a hybrid action space which consists of primitive robot actions and temporally extended nominal options that transfer control to one of the model-based controllers.
During exploration, when the agent takes a nominal action at a state, it verifies in simulation whether or not the task can be solved reliably by transferring control to the nominal controllers from that state.
The verification is then used as a binary reward signal for the recovery policy.
RecoveryChaining not only learns how to recover from the immediate source of failure but also which nominal controller to recover to.
Furthermore. we propose \textbf{Lazy RecoveryChaining} that improves sample efficiency by continually training precise binary classifiers for reward generation to avoid expensive simulation of nominal options in previously visited states.

\begin{figure*}[th]
    \centering
    \includegraphics[width=1\linewidth]{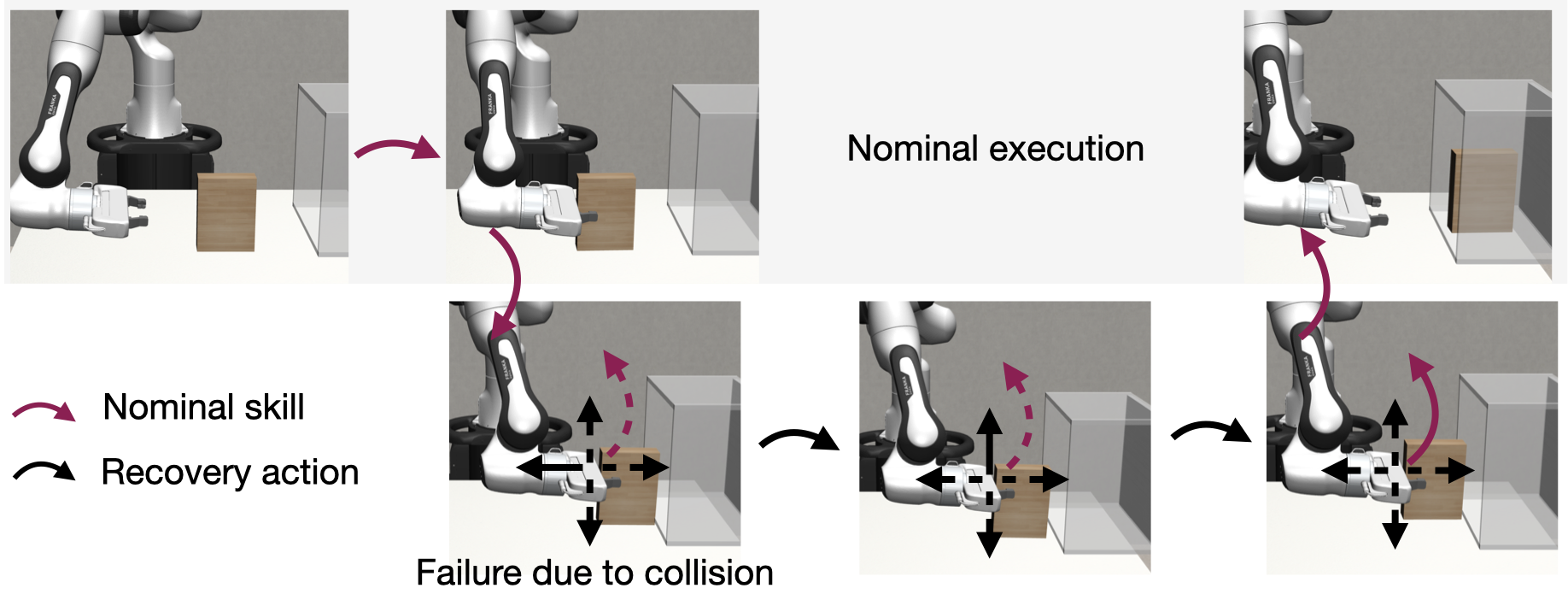}
    \caption{
    We propose an approach to learn robust recovery behaviors on top of given nominal controllers using reinforcement learning that works even with sparse rewards.
    Here, the robot is trying to place a box on a shelf but accidentally collides with the shelf due to an imprecise grasp.
    Using our approach, the robot learns a recovery policy from the failure state in a \emph{hybrid} action space consisting of primitive robot actions and temporally extended nominal options that trigger a sub-sequence of the nominal controllers.
    The recovery policy is trained to quickly take the robot to the precondition of one of the nominal controllers so that it can transfer control to the nominal controllers to complete the task.
    Solid arrows indicate actions taken by the robot and dashed arrows other available actions.
    }
    \label{fig:approach_overview}
\end{figure*}
We evaluate our approach in three multi-step manipulation tasks of pick-place, shelf, and cluttered-shelf. The results show that our approach is able to learn significantly more robust recoveries than prior methods.
In some of the shelf domain scenarios, the robot is able to learn to leverage contact with the environment to reduce uncertainty, adjust for slip, and avoid further collisions.
We also show that our approach is suitable for sim-to-real by transferring the learned skills to a physical Mitsubishi Electric Assista arm without the need of any real-world fine-tuning (see figure~\ref{fig:robot_execution}).

\section{Related Work}

\subsection{Recovery Learning}
Recent works have explored the idea of learning recovery policies using offline datasets for safe exploration~\cite{thananjeyan2021recovery,wilcox2022ls3} and for recovery from execution failures~\cite{vats2023efficient,reichlin2022back}.
For example, \cite{wilcox2022ls3} learn a safe set using an offline dataset and \cite{vats2023efficient} learn skill preconditions by executing the skills from different initial states.
The learned preconditions (or safe sets) are then used as the goal for recovery learning.
This approach is  highly sensitive to the quality of the learned preconditions and is pessimistic as it tries to stay close to the offline dataset.
To address these issues, our approach uses online RL with temporally extended actions to better explore the state space and discover robust policies.

Another common approach for learning reactive policies is to learn from human demonstrations~\cite{chernova2014robot,chi2023diffusion}.
However, these policies are prone to failure if the robot visits out-of-distribution states during execution.
Online data collection~\cite{ross2011reduction} by an expert human is often required to learn recoveries which makes this approach quite expensive.
By contrast, our approach does not rely on human demonstrations.
\cite{wang2022tli} propose to recover by modulating dynamical systems learned from segmented demonstrations.
However, this assumes accurate detection of manually specified modes which we do not require.

\subsection{Hierarchical Reinforcement Learning}
Hierarchical reinforcement learning (HRL)~\cite{pateria2021hierarchical} is an approach for solving long-horizon decision making problems.
HRL focuses on decomposing a problem into smaller sub-tasks to enable easier policy learning and better generalization.
This decomposition allows decisions to be made at higher levels of abstraction without having to deal with the low-level details.
Skill chaining~\cite{konidaris2009skill} is a popular approach to discover skills that allow an agent to solve any task in the state space. A major challenge in skill chaining is to learn reliable initiation sets. \cite{bagaria2021robustly} address this by proposing a combination of pessimistic and optimistic classifiers.
However, such approaches have primarily been evaluated in navigation domains and transfer to high-dimensional domains such as manipulation remains an open area of research.
Our recovery learning approach utilizes the HRL framework, wherein recovery policies from failures are learned to connect them to the nominal policies.
Our approach is inspired by previous works which show that manipulation policies can be learned more efficiently by using structured action spaces~\cite{sharma2020learning,nasiriany2022augmenting,rosen2022role} such as, object-centric controllers and parameterized primitives.




\section{Background}
\textbf{Markov Decision Process (MDP).}
A Markov Decision Process (MDP)~\cite{sutton2018reinforcement} is defined by the tuple $({S}, {A}, T, R, \gamma, \mu)$ where $\mathcal{S}$ is the state space, ${A}$ is the action space, $T$ is the transition function, $R$ is the reward function, $\gamma$ is the discount factor and $\mu$ is the initial state distribution.
The MDP framework has been used extensively to model tasks in manipulation for planning and reinforcement learning.
A key assumption made by MDPs is that the system state is always accurately known.
This may not be true for some state variables, for example, due to errors in pose estimation of objects or some objects being out of the sensor's field of view.
The Mixed Observable MDP (MOMDP)~\cite{pomdp2010ong} accounts for this uncertainty by factoring the state into two types of state variables: fully observable variables represented by the variable  $x$ and partially observable variables $y$.
The tuple $(x, y)$ fully specifies the system state.
For example, a robot's end-effector pose is usually known quite accurately and hence should belong to $x$, while poses of occluded objects would belong to $y$.




\textbf{Options Framework.}
We model each robot skill as an option as per the options framework~\cite{sutton1999between}. Each option consists of three components:  (a) a robot \emph{control policy} $\pi$ (b) an \emph{initiation set} $\mathcal{I}$, also sometimes called a \emph{precondition}, which defines the states from which the option can be executed and (c) a \emph{termination condition} $\beta$  which defines the states in which the option must terminate.
\rebuttal{Once an option terminates, the robot may choose to initiate another valid option. This decision is made by a separate high-level policy over options which treats the available options as its action space.
}

\textbf{Skill Chaining.}
Skill chaining (SC)~\cite{konidaris2009skill} is a popular approach to solve long horizon RL problems by decomposing it into shorter sub-problems.
A sequence of options is learned backwards from the goal, such that in each iteration a new option is learned to reach the precondition of the previously learned options.
The precondition is usually learned using binary classification and describes the states from which the option policy can be successfully executed.
During the initiation period of an option it is executed from different states in the environment to collect data for training the precondition.
The precondition classifier is then frozen and used to generate success or failure rewards for the option that is trying to reach it.

\section{Problem Statement}
\begin{figure}[t]
    \centering
    \includegraphics[width=1\columnwidth]{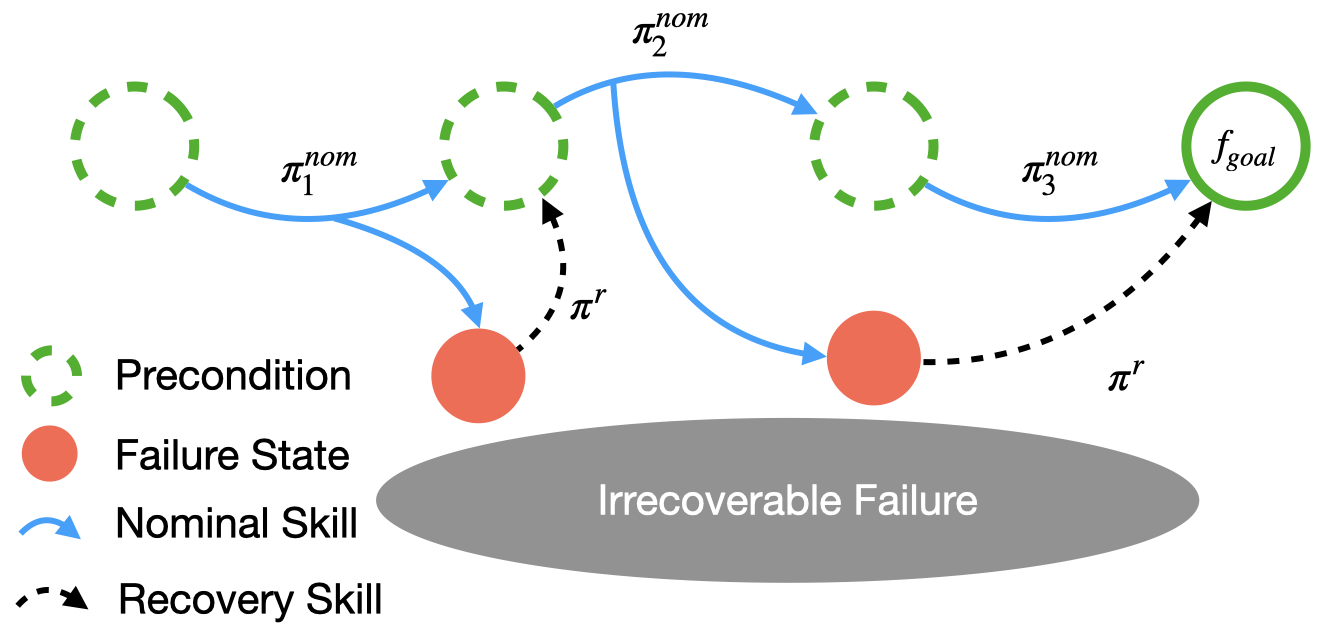}
    \caption{Representation of a sequence of nominal policies that solve a task specified by a binary function $f_{goal}$.
    \rebuttal{Due to model inaccuracies and stochastic dynamics, the system may deviate from the nominal plan. A failure detector is used to stop the robot before it encounters an irrecoverable failure. However, this state could be outside the preconditions of the nominal policies.
    Hence, a new recovery policy $\pi^r$ is learned to take the system back on the nominal plan.
    }
    }
    \label{fig:hrl_incoming}
\end{figure}

Consider a long-horizon manipulation task defined by a distribution of start states and a binary goal function $f_{goal}: S \rightarrow \{0,1\}$.
\rebuttal{
As shown in figure~\ref{fig:hrl_incoming}, we assume we are given a \emph{nominal plan} $\xi^{nom} = (\pi^{nom}_1,\ldots,\pi^{nom}_k)$ made up of $k$ nominal controllers (or policies) that is capable of solving the task with non-zero probability. The nominal plan sequentially executes each nominal policy until the policy's termination condition is met. We use a fixed maximum number of steps as the termination condition for all the policies in our experiments unless the goal or failure condition is met.
Due to state and actuation uncertainty or model inaccuracy, the system may deviate from the intended plan. We assume that we have a failure detector that can detect impending irrecoverable failures and hence ensure that the robot is always in a recoverable state. However, the actual nominal initiation sets are unknown.
}
Our goal is to robustify the system by efficiently learning a separate \emph{recovery policy} that allows the robot to complete the task after failure detection.
In the following section, we describe our proposed approach and present solutions to some of the key challenges associated with this learning problem.

\section{Approach}
We model the system as a MOMDP~\cite{pomdp2010ong}, wherein the robot maintains an estimate $\hat{s} := (x, \hat{y})$ of the true state  $s$ and acts based on $(\hat{s}, o)$, where $o \in O$ corresponds to sensory observations such as proprioception.
Our approach, visualized in fig.~\ref{fig:approach_overview}, involves two steps: (1) \textbf{Failure Discovery.} Nominal policies are executed under various conditions in simulation to induce and record failures.
We leverage privileged information in simulation to record both the true state $s$ and the corresponding observations associated with a failure.
This allows us to directly reset to the failure in simulation during recovery learning.
(2) \textbf{Recovery Learning.}
Recovery policies are learned in simulation to handle the failures collected in the previous step using reinforcement learning.
\rebuttal{
At deployment, the robot executes the nominal plan and switches to the recovery policy if a failure is detected.
}

\subsection{Failure Discovery}
\rebuttal{
\textbf{Failure Detection.}
We utilize a failure detector that monitors the execution of the system and raises a flag \texttt{fail-condition} if unsafe or unexpected conditions are met, for example, high end-effector forces, dropping an object, or slip. We assume that the failure detector can prevent the robot from encountering irrecoverable failures.
While we hand-design the failure detectors in our experiments, prior works have shown how they can be learned from data, e.g., Associative Skill Memories~\cite{pastor2012towards} flag deviation of sensor measurements from previously collected successful trajectories as a failure and ~\cite{vats2023efficient} train a failure classifier using successful and failed trials.
We note that failure detection is an active area of research and is not guaranteed for all types of tasks.
}

\textbf{Failure Discovery.} We discover and record potential failures $\mathcal{D}_{fail}$ of the nominal policies by executing them from various initial conditions.
Once a failure is detected, execution is terminated and the resulting state $s = (x,y)$ is recorded.
Though, the robot may not have access to the true world state at test time, we can do so during training either in simulation or with extra sensors in the physical world.
This is straightforward in simulation but \rebuttal{can be done with extra sensing on a real system for some tasks, for example, using AprilTag markers to measure object positions. In our shelf experiments, box position is partially observable, i.e., $y = pos_{box}$ while $y = \{\}$ in pick-place  since the environment is fully observable.}

\subsection{Recovery Learning}
Our goal is to learn a policy that reliably takes the robot to the precondition of one of the nominal policies from all the failures $\mathcal{D}_{fail}$.
For example, if the book in a robot's hand slips while putting it on a shelf, recovery should regrasp the book such that the place controller can be executed.
Since the true preconditions $\mathcal{I}_i^{nom}$ are usually unknown, prior works~\cite{thananjeyan2021recovery,wilcox2022ls3,reichlin2022back} in recovery learning first estimate the nominal precondition \rebuttal{$\hat{\mathcal{I}}_i^{nom}$} offline.
The RL agent then computes actions to maximize the estimated precondition \rebuttal{$r(s,a) = \hat{\mathcal{I}}_i^{nom}(s)$}.
However, accurate estimation of preconditions is highly dependent on the quality of the offline dataset making these approaches brittle and pessimistic.

\rebuttal{To address these drawbacks, we propose an online RL approach that does not rely on known preconditions.
Instead of using learned preconditions, our main observation is that we can compute a monte-carlo estimate of the precondition of a nominal controller $\pi^{nom}_i$ by executing the nominal plan suffix $o_i^{nom} := (\pi^{nom}_i,\ldots,\pi^{nom}_k)$ at the query state $s$.  If the plan terminates in a state $s'$ inside the goal, then the precondition is satisfied at $s$, i.e., the precondition can be estimated as $\mathcal{I}^{mc}_i(s) = f_{goal}(s')$.
}
A single monte-carlo simulation is sufficient in deterministic MDPs but an average of multiple simulations may be required in stochastic domains. We use only a single simulation in our experiments.

\textbf{A Straw-man Approach.}
A straw-man RL approach to leverage this observation would be to replace the learned $\mathcal{I}^{nom}_i$ with $\mathcal{I}^{mc}_i$.
However, this is highly inefficient as the environment would have to execute long sequences of nominal controllers, potentially multiple times, in every step to compute the reward.
Ideally the environment need not have to repeatedly compute the monte-carlo estimates from states well outside the precondition.

\textbf{RecoveryChaining.}
Our key insight is \emph{to let the RL agent decide when to estimate the precondition}.
We provide the RL agent with temporally extended \emph{nominal} options $o^{nom}_i$ 
 which simulate the transfer of control to the nominal policy $\pi^{nom}_i$.
The sequence of nominal policies $o^{nom}_i = (\pi^{nom}_i,\ldots,\pi^{nom}_k)$ is then executed from that state and the resulting task success is used as a reward.
We choose to make the nominal options terminal so that the corresponding reward is independent of the policy being learned and hence is stationary.
Monte-carlo simulations are computationally more expensive than querying a learned precondition but provide a more reliable reward estimate in our experience.

\begin{figure}[t]
    \centering
    \includegraphics[width=1\columnwidth]{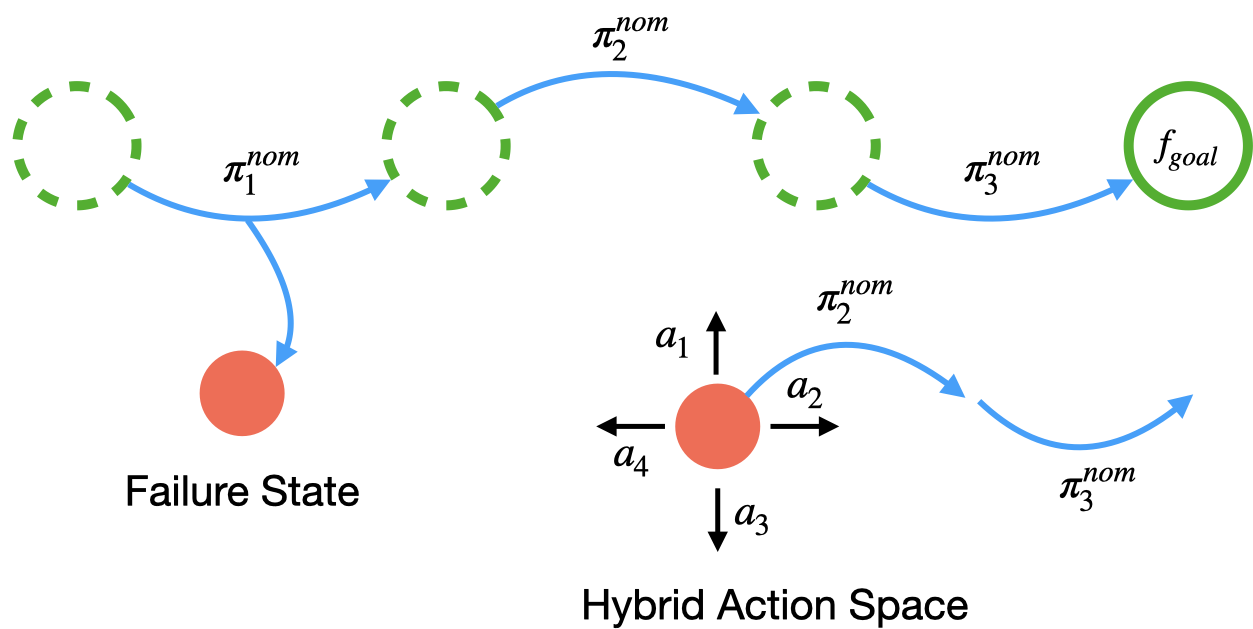}
    \caption{
    We use a hybrid action space for reinforcement learning.
    It consists of both primitive robot actions and nominal options that transfer control to a sequence of nominal policies that can take it to the goal if applied successfully.
    }
    \label{fig:rl_action_space}
\end{figure}

\textbf{RecoveryChaining MDP.}
Let $A$ be the original action space of the agent consisting of primitive actions and $S$ be the state space. \rebuttal{$S$ contains an absorbing goal state $s_g$ and an absorbing failure state $s_f$. If the robot satisfies $f_{goal}$, then it transitions to $s_g$ with  a reward of $1$. If \texttt{fail-condition} is triggered instead, then it transitions to $s_f$ with a reward of $0$.}
The RecoveryChaining MDP is defined by the tuple $(S^{rc}, A^{rc}, T^{rc}, r^{rc}, \gamma, \mu^{rc})$ where
\begin{itemize}
    \item \rebuttal{$S^{rc} = S \cup \{s_d\}$, where $s_d$ is a new absorbing state.}
    \item $A^{rc}$ is a hybrid action space $A \cup \{o^{nom}_1,\ldots,o^{nom}_k\}$ consisting of primitive actions and terminal nominal options that transfer control to the nominal policies.
    \item \rebuttal{The agent transitions to $s_d$ after executing $o^{nom}_i \ \forall \ i$ if it does not transition to the goal.}
    \item \rebuttal{$r^{rc}(s \in S) = f_{goal}(s)$, $r^{rc}(s_d) = 0$.}
    \item $\mu^{rc} = \mathcal{D}_{fail}$
\end{itemize}
The hybrid action space is visulialized in figure~\ref{fig:rl_action_space}.
Intuitively, when the agent is far away from the precondition $\mathcal{I}^{nom}_i$ and executes $o^{nom}_i$, it gets no reward.
After a few trials and errors, the agent identifies states that lie outside the precondition and stops executing the nominal option from those states.
On the other hand, as the agent gets closer to the precondition, it receives higher rewards upon executing $o^{nom}_i$ and implicitly learns that it is inside the desired precondition.

This approach has three key advantages:
\begin{enumerate}
    \item The agent implicitly learns the nominal preconditions through trial and error and stops computing monte-carlo estimates for nominal controllers that are not applicable.
    \item The ability to try all the nominal policies allows the agent to potentially discover novel ways to reuse them.
    \item The agent is not obligated to recover to any precondition.
    If the correct recovery is to completely avoid the nominal policies then it can discover such strategies.
\end{enumerate}


\subsection{Lazy RecoveryChaining} Simulation of temporally extended nominal options is the most computationally expensive step in RecoveryChaining. In fact, as recovery policy improves, it spends an increasingly greater chunk of its time evaluating these nominal actions in known good states because these yield positive rewards. However, this is inefficient, as we would rather explore uncertain and previously unvisited regions. To address this, we propose to continually learn regions in which the nominal option is highly likely to succeed using a conservative binary classifier, $\alpha$. Specifically, for every nominal option $o^{nom}_i$, we train a precise binary classifier $\alpha_i$  using data from its previous mote-carlo rollouts. If the RL agent selects a nominal option from  a state where the classifier is confident of success, we lazily assign a positive reward instead of executing the full rollout.

To mitigate approximation error in classification and hence reward generation, we utilize two strategies. First, we ensure low false positive rates by training binary classifiers with high precision. In our experiments, we use XGBoost~\cite{chen2016xgboost} to learn probabilistic classifiers and select thresholds that guarantee high precision. We continually retrain these classifiers using on-policy data and apply them only when their precision meets a stringent threshold ($\geq 0.95)$. 
Second, to mitigate bias and maintain a balanced dataset, we randomly perform monte-carlo rollouts with a small ($20\%$) probability even when the classifiers are confident.


\section{Experiments} \label{sec:experiments} 
\begin{figure*}[t]
    \centering
    \includegraphics[width=0.95\linewidth]{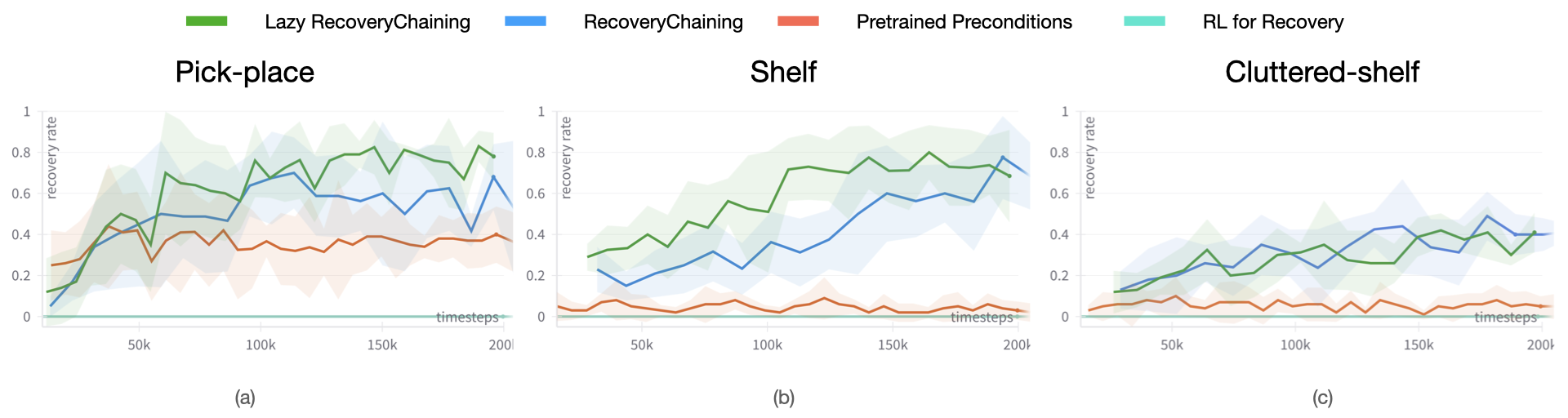}
    \caption{Comparison of the learning curves of RecoveryChaining (RC), Lazy RC, Pretrained Preconditions (PP) and RL for Recovery (RLR) in pick-place, shelf and cluttered-shelf domains.
    RC and Lazy RC make consistent progress in learning, with Lazy RC learning faster in 2/3 domains. PP hits a local optimum early in training and is not able to further improve its policy as it is limited by a pretrained reward model. PP does quite poorly on the shelf task due to its partially observable nature. RLR makes no progress in any of the tasks.
    Results are averaged over 5 different seeds.
    }
    \label{fig:learning_curve}
\end{figure*}

We evaluate our proposed approach in three challenging multi-step manipulation environments with sparse rewards.
The purpose of these experiments is to understand (1) whether our approach can learn robust and composable recoveries and (2) how well the hybrid approach of combining model-based and model-free policies works.
\rebuttal{While a model-based planner can be used to compute nominal plans, we design them manually in our experiments as planning is not the focus of our work.}

\textbf{Baselines.}
We compare RecoveryChaining (RC) and Lazy RC with three baselines.
(a) \emph{Nominal:} completes the task using just the model-based controllers;
(b) \emph{Pretrained Preconditions (PP):} learns a recovery policy using primitive actions to reach preconditions learned from offline data;
\rebuttal{(c) \emph{RL for Recovery (RLR):} learns the recovery policy using standard RL with primitive actions.}
All the methods use a \emph{sparse reward function}.

\textbf{RL Training.}
We use a set of discrete action primitives to learn all the policies for easier sim-to-real transfer.
Each action primitive is defined in the robot's end-effector frame and operates only on a single dimension.
Translation action primitives move the end-effector by $\pm 2$ cm along $x$, $y$ or $z$ axis and rotational primitives apply a roll, pitch or yaw of $\pm \pi/2$.
We train all RL approaches for $200$K timesteps using Proximal Policy Optimization (PPO)~\cite{schulman2017proximal} from stable-baselines~\cite{raffin2019stable}.
We run all the methods using $5$ different seeds and report the average.



\subsection{Pick-Place Domain}
\begin{figure}[]
    \centering
    \includegraphics[width=0.8\columnwidth]{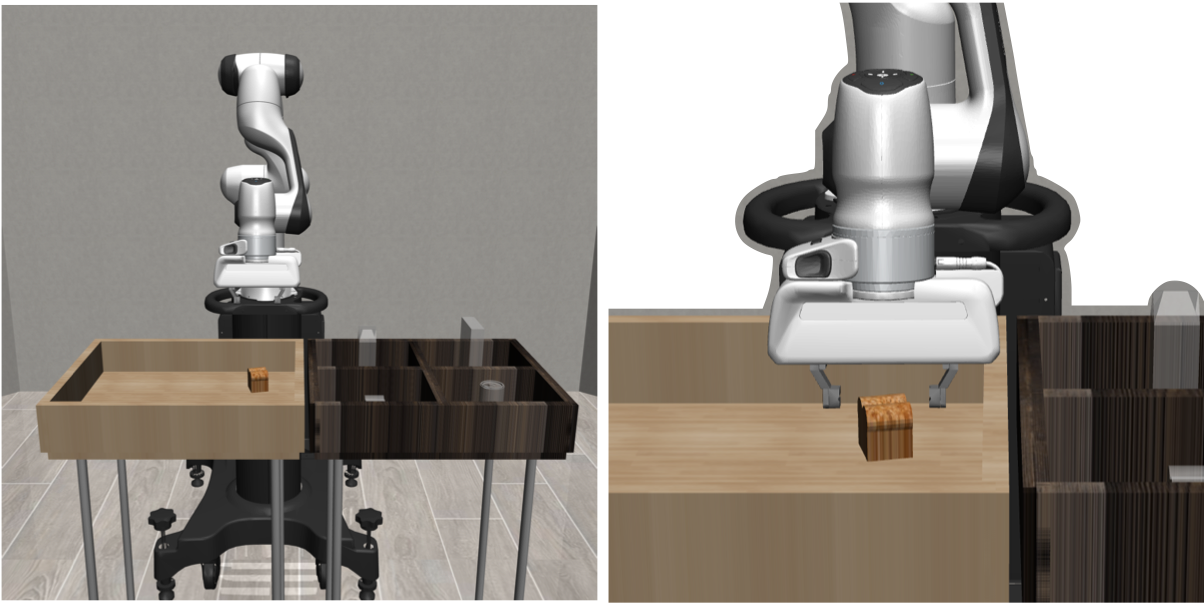}
    \caption{The pick-place task requires the robot to pick a small bread from the source bin and place it in the target bin.
    The nominal controllers do not account for the sides of the bin because of which the end-effector collides with them  when the bread is close to the walls.
    One such situation is shown in the right figure.}
    \label{fig:pick_place_env}
\end{figure}
Our first domain is the pick-place task from robosuite~\cite{robosuite2020}.
To solve this task, the robot needs to pick a small loaf of bread from the source bin and place it in the target bin.
The robot gets a 46 dimensional observation consisting of object poses, end-effector pose, etc.
The bread is initialized in a different location in the source bin in every episode.

\textbf{Nominal Skills.}
We designed four nominal controllers: \textsc{GoToGrasp} skill takes the robot to a pre-grasp pose over the object, \textsc{Pick} skill picks up the object, \textsc{GoToGoal} moves the robot to the drop location and \textsc{Place} skill places the object at the drop location.
Each controller is implemented as a state machine that selects a target pose for the robot end-effector in the Cartesian space based on its current state.
The target end-effector pose is then achieved using task-space impedance control with fixed impedance.
The nominal skills achieve a success rate of $70\%$ on their own in this task.

\textbf{Failures.}
Most of the failures in this task occur when the initial location of the  bread is close to one of the walls.
The robot needs to reach inside the bin to grasp the object due to its small size.
This causes the end-effector to collide with the wall leading to a failure that the nominal skills aren't capable of handling.
Collision is detected using a threshold on the end-effector forces.
We collect a total of $100$ failures for learning recovery.

\textbf{Recovery.}
The learned policy finds two different strategies to recover from these failures.
The first strategy rotates the end-effector along the $z$ axis so that it does not collide with the wall when the robot reaches inside the bin.
The second, and perhaps more interesting, strategy uses the gripper fingers to push the object away from the walls before picking it up. 
RC uses a combination of these two strategies to improve the  success rate from $70\%$ to $90\%$.
On the other hand, PP quickly plateaus and converges to a locally optimal policy (fig~\ref{fig:learning_curve}).
It uses preconditions learned from around 300 nominal trajectories as the reward function.
RLR is not able to learn the recovery at all due to sparse reward.
We summarize the success rates in table~\ref{table:results1}.

\subsection{Shelf Domain}
Our second domain  is a shelf environment with state uncertainty also implemented using robosuite.
In this task, (fig.~\ref{fig:novel_push}) the robot needs to pick up a box from a table and place it inside a shelf in an upright position.
The robot observes a noisy estimate of the position of the box, where, the noise is sampled from a zero-mean Gaussian distribution with standard deviation of $1$ and $2$ cm along the $y$ and $z$ axes, respectively.
We also provide the robot the number of actions taken so far as an observation.
This allows it to learn open-loop policies if needed.
The dimensions of the shelf and the box and the position of the shelf are sampled randomly in every episode.

\textbf{Nominal Skills.} We designed three nominal skills for the task assuming that the observations are accurate.
\textsc{Pick} skill: the robot goes to the observed position of the box, closes the gripper, and picks up the box; \textsc{Move} skill: the robot moves to a pre-placement position conditioned on the given position of the shelf; \textsc{Place} skill: the robot places the box on the shelf and retracts.
These nominal skills can complete the task reliably if the state estimates are accurate but can fail when the state estimates are wrong.
All the nominal skills control the robot using task-space impedance control with fixed impedance.


\textbf{Failures.}
We collect failures by executing the nominal policies in simulation under state uncertainty.
The failure conditions are given by the end-effector forces $F_e$  exceeding a predefined threshold.
The nominal controllers assume perfect pose of the box.
However, the robot may grasp the box with an offset due to a wrong position estimate.
This leads to mainly two types of failures: (1) \emph{collision}: robot collides with the shelf or the table (2) \emph{collision-slip}: collision with the shelf leads to in-hand rotation of the object if there is a delay in stopping the robot.

\textbf{Recovery.}
The recovery policy learns to move up and inside the shelf before switching to the nominal skill because most collisions happen below the center of mass of the object.
We found that providing the number of past actions was crucial to learn a recovery in this task because of unreliable state estimates.
In our ablation studies with different amounts of state uncertainty, we found that the policy tends to be more conservative and learns relatively open-loop policies under high uncertainty.
Overall, RC did significantly better than PP on this task by improving task success from $50.8\%$ to $83.2\%$. Lazy RC achieved a final recovery rate similar to RC but was more sample efficient.
RC was not able to recover from failures involving significant in-hand rotation of the box as it was not provided with its orientation observation.
Additional sensing, for example, slip detection using a tactile sensor can further improve the recovery policy.
The reward model for PP  was learned using 232 nominal trajectories.

\begin{table}
\centering
    \begin{tabular}{llllc}\toprule
                           & Nom &  \textbf{Nom + RC} & Nom + PP & Nom + RLR      \\ \midrule
                            
        Pick-place &  70 & \textbf{90} & 76  &  70 \\
        Shelf     & 51 & \textbf{83} & 56 & 52 \\
        Cluttered-shelf & 38 & \textbf{57} & 43 & 41 \\
      \end{tabular}
    \caption{Comparison of the overall success rate (\%).
    RecoveryChaining (RC) significantly robustifies the nominal controllers and is the best performing method in all the domains. The improvement is more pronounced in the shelf domain as it has state uncertainty which makes learning more challenging.
    }
\label{table:results1}
\end{table}

\subsection{Cluttered Shelf Domain}
\begin{figure}[th]
    \centering
    \includegraphics[width=0.40\columnwidth]{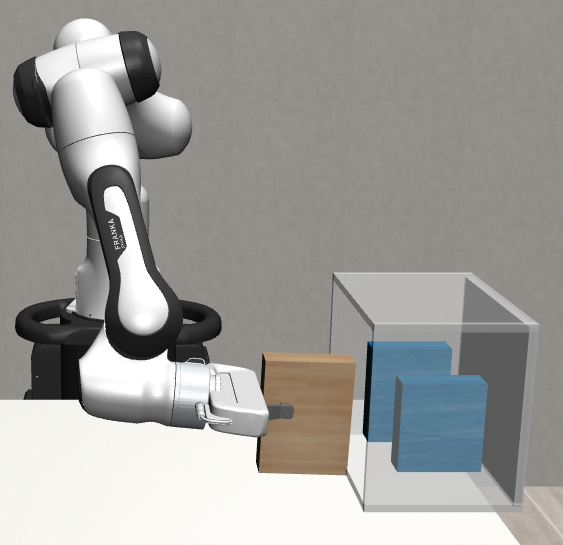}
    \includegraphics[width=0.36\columnwidth]{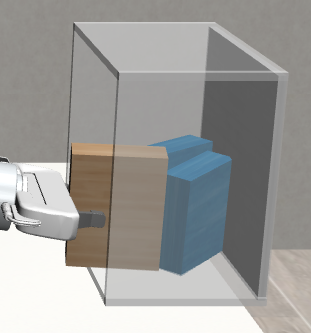}
    \caption{\emph{(left)} Cluttered shelf domain. The robot needs to place a box on a cluttered shelf with two objects.
    In addition to avoiding collision, successful task completion requires the robot to avoid rotating the objects on the shelf.
    \emph{(right)} Failure state. The robot collides and rotates the objects during execution leading to a failure.}
    \label{fig:clutter_shelf_env}
\end{figure}

We consider a more complex version of the shelf domain with two objects randomly placed on the shelf.
The robot needs to avoid them while putting the box on the shelf.
We use the same nominal skills used in the shelf domain.

\textbf{Failures.}
In addition to monitoring end-effector forces for collision detection, we also use vision-based failure detection to avoid toppling objects on the shelf.
The failure detector is triggered if the robot topples or rotates any object by more than a predefined threshold.
Such failures can be detected in the real world by using object detectors and pose estimation while we use privileged state information in simulation.
We show an example of a  failure in figure~\ref{fig:clutter_shelf_env}.

\textbf{Recovery.}
We compare the learning curves of RC with PP in figure~\ref{fig:learning_curve}.
RC learns a significantly more reliable recovery skill than that learned by PP.
However, we observe a drop in performance compared with that on the simpler shelf task.
We believe performance can be improved with longer training, a dense reward function and a better action space.

\subsection{Analysis of Learned Recoveries}
\begin{figure}[t]
    \centering
    \includegraphics[width=0.85\columnwidth]{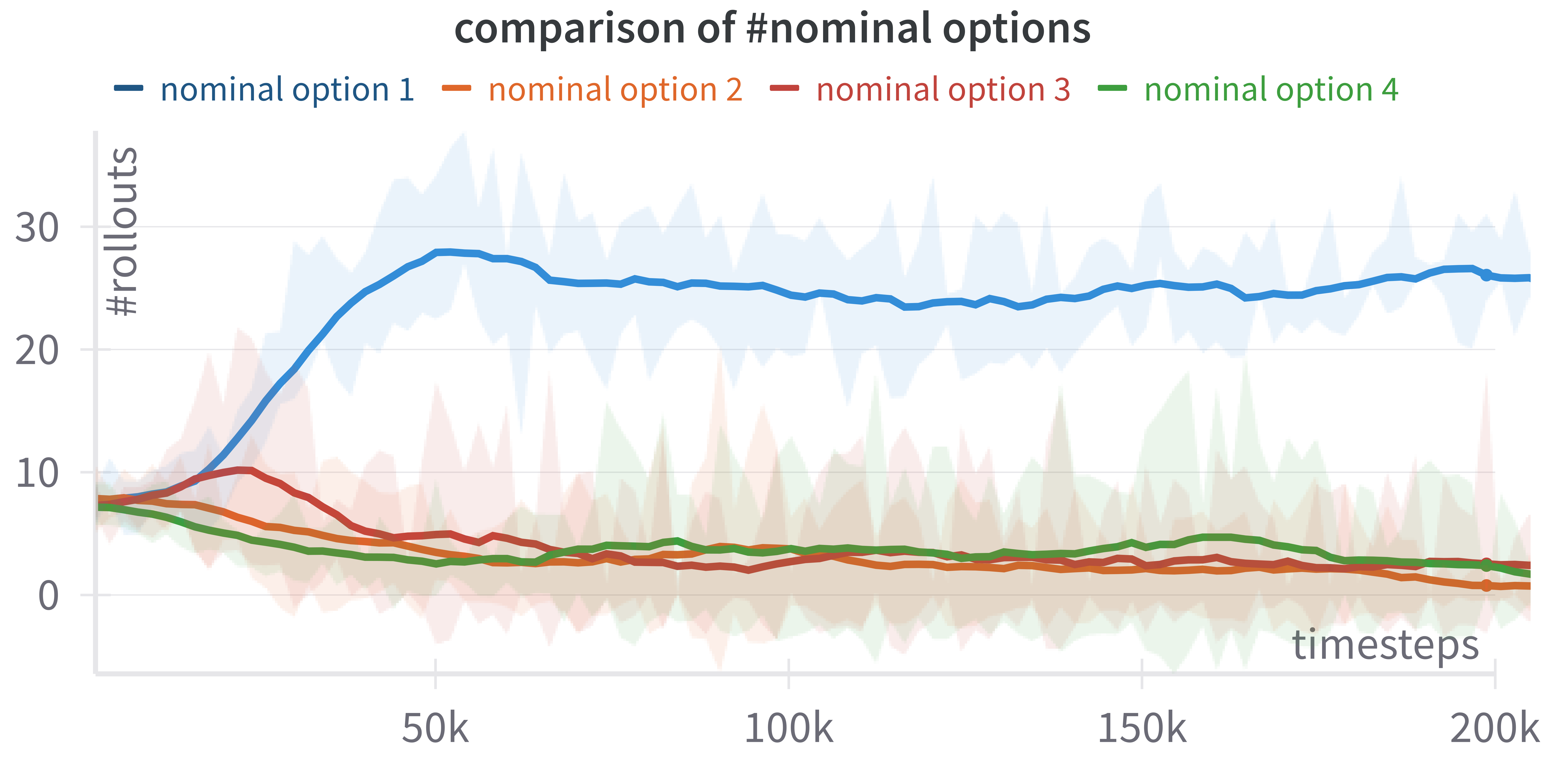}
    \caption{A comparison of the number of different nominal options taken by the agent in the pick-place task in every round of exploration consisting of 120 actions.
    The agent explores all the nominal options initially but quickly identifies and commits to the best nominal controller to recover to.
   }
    \label{fig:comparison_nom_actions}
\end{figure}

\textbf{RC can identify the best nominal controller to recover to.}
We do not provide the agent any prior knowledge about where it should recover to and which nominal controller it should switch to.
To understand the exploration behavior of RC we plot the number of times the agent chooses each nominal option in a round of exploration in figure~\ref{fig:comparison_nom_actions}.
Each round consists of 120 actions.
We observe that the agent initially explores all the nominal controllers but quickly identifies the most suitable one and commits to recovering to its precondition.
The RC policy \emph{implicitly} learns the preconditions of all the nominal controllers through trial and error and avoids trying to switch to a nominal controller at states in which it is unlikely to succeed.

\begin{figure}[t]
    \centering
    \includegraphics[width=0.24\columnwidth]{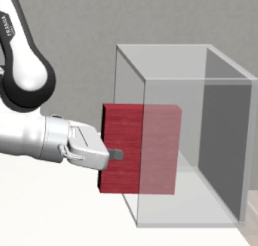}
    \includegraphics[width=0.24\columnwidth]{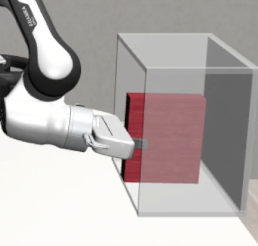}
    \includegraphics[width=0.24\columnwidth]{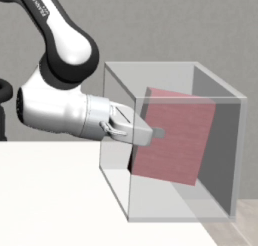}
    \includegraphics[width=0.24\columnwidth]{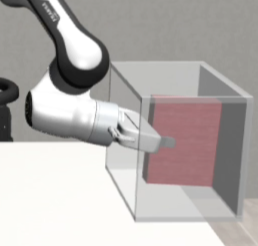}
    \caption{While trying the nominal controllers from different states during exploration, the agent discovers a novel application of the \textsc{place} controller.
    \emph{(top)} The \textsc{place} skill was designed to gently place the box assuming it is upright.
     \emph{(bottom)} To fix the slip due to a prior collision, RC learns to move deeper inside the shelf than nominal execution before switching to the \textsc{place} skill.
    This allows the robot to fix the orientation of the box by pushing against the back of the shelf to ensure stable placement.}
    \label{fig:novel_push}
\end{figure}

\textbf{RC can reuse the nominal controllers in novel ways.}
Prior approaches that use pretrained preconditions, like skill chaining, provide a pessimistic bias to the RL agent by freezing the preconditions after the initiation period.
This prevents the agent from discovering novel ways to use the nominal controllers that may be quite different from the trajectories used to train the precondition.
Sometimes, the robot needs to use the nominal controllers in novel ways not seen during nominal execution.
We describe one such novel reuse of the \textsc{place} skill discovered by our agent for the shelf task in figure~\ref{fig:novel_push}.
In this example, the box undergoes significant in-hand rotation due to a collision with the shelf below its center of mass.
The robot needs to re-grasp the box before placing it on the shelf as it may fall over otherwise.
The RC agent discovers that switching to the nominal \textsc{place} skill from  inside the shelf fixes the in-hand slip by aligning the box with the back of the shelf.
This behavior is well outside the distribution of states visited by the nominal skills as the \textsc{place} controller is designed to be triggered outside the shelf.

\subsection{Transfer to a Physical Robot}
We transfer the nominal and recovery policies to a real robot.
Our real-world setup includes a 6-DoF Mitsubishi Electric Assista robot arm with a WSG-32 parallel-jaw gripper and a Mitsubishi F/T sensor 1F-FS001-W200 mounted on the wrist of the robot.
We evaluate the feasibility of sim-to-real transfer of recoveries learned using our approach.
We train the recovery in a simulated Assista robot with a box and then evaluate it on a real Assista with one box and two unseen objects- a mustard bottle and a can.
Failures are induced on the real robot by providing incorrect position estimates of the shelf.
Similar to the simulation, we implement a collision detector on the real robot by using observations from the F/T sensor on the robot.
The robot was able to recover from both collision and slip (fig.~\ref{fig:robot_execution}) by using our recovery.
The policy generalized well to the mustard bottle but performed slightly worse on the can because it is smaller than the box that was used for training and its curved surface is more prone to slip.

\begin{table}[ht]
\centering
    \begin{tabular}{lc}\toprule
                             & Recovery Rate (\%)        \\ \midrule
                             
        Box                  & 100 (5/5)   \\
        Mustard bottle       & 100 (5/5)   \\
        Can                  & 80  (4/5)   \\
      \end{tabular}
    \caption{Summary of real-robot experiments on a real robot.
    Our approach generalizes to a mustard bottle and a can despite having been trained only on a box.}
\label{table:results3}
\end{table}


\section{Conclusion} \label{sec:conclusion}
We propose a hierarchical reinforcement learning approach to learn model-free recovery policies for robustifying nominal model-based controllers.
Our approach, called RecoveryChaining, uses a hybrid action space to efficiently learn robust recovery policies that can be chained with model-based controllers.
The action space contains temporally extended nominal options that transfer control to a specific nominal controller.
These nominal options reduce the effective horizon of the task and enable recovery learning for multi-step manipulation.
We evaluate our approach in three challenging domains and find that our approach can significantly improve task success using just a sparse reward.
We also transfer a recovery trained in simulation to a physical robot to demonstrate the feasibility of sim-to-real transfer.

A major limitation of our method is its reliance on a physics-based simulator for recovery learning.
This limits its applicability to tasks that can be modeled well in simulation and introduces the challenge of sim-to-real transfer.
Second, we assume that there exists an initiation set from which the nominal policies can be reliably executed and that this set can be reached from failures using a local corrective policy. Our method will not be effective if the nominal policies are unreliable everywhere since the recovery policy will not find any good state to switch to the nominal plan.
Finally, more research is needed in efficiently learning recoveries in partially observable environments.

\balance


\bibliographystyle{IEEEtran}
\bibliography{references}

\begin{thebibliography}{10}
\providecommand{\url}[1]{#1}
\csname url@rmstyle\endcsname
\providecommand{\newblock}{\relax}
\providecommand{\bibinfo}[2]{#2}
\providecommand\BIBentrySTDinterwordspacing{\spaceskip=0pt\relax}
\providecommand\BIBentryALTinterwordstretchfactor{4}
\providecommand\BIBentryALTinterwordspacing{\spaceskip=\fontdimen2\font plus
\BIBentryALTinterwordstretchfactor\fontdimen3\font minus \fontdimen4\font\relax}
\providecommand\BIBforeignlanguage[2]{{%
\expandafter\ifx\csname l@#1\endcsname\relax
\typeout{** WARNING: IEEEtran.bst: No hyphenation pattern has been}%
\typeout{** loaded for the language `#1'. Using the pattern for}%
\typeout{** the default language instead.}%
\else
\language=\csname l@#1\endcsname
\fi
#2}}

\bibitem{mason2001mechanics}
M.~T. Mason, \emph{Mechanics of robotic manipulation}.\hskip 1em plus 0.5em minus 0.4em\relax MIT press, 2001.

\bibitem{kroemer2021review}
O.~Kroemer, S.~Niekum, and G.~Konidaris, ``A review of robot learning for manipulation: Challenges, representations, and algorithms,'' \emph{The Journal of Machine Learning Research}, vol.~22, no.~1, pp. 1395--1476, 2021.

\bibitem{lagrassa2020learning}
A.~Lagrassa, S.~Lee, and O.~Kroemer, ``Learning skills to patch plans based on inaccurate models,'' in \emph{2020 IEEE/RSJ International Conference on Intelligent Robots and Systems (IROS)}.\hskip 1em plus 0.5em minus 0.4em\relax IEEE, 2020, pp. 9441--9448.

\bibitem{ebert2018robustness}
F.~Ebert, S.~Dasari, A.~X. Lee, S.~Levine, and C.~Finn, ``Robustness via retrying: Closed-loop robotic manipulation with self-supervised learning,'' in \emph{Conference on Robot Learning}.\hskip 1em plus 0.5em minus 0.4em\relax PMLR, 2018, pp. 983--993.

\bibitem{wang2019robust}
A.~S. Wang and O.~Kroemer, ``{Learning robust manipulation strategies with multimodal state transition models and recovery heuristics},'' \emph{Proceedings - IEEE International Conference on Robotics and Automation}, vol. 2019-May, pp. 1309--1315, 2019.

\bibitem{sundaresan2021untangling}
P.~Sundaresan, J.~Grannen, B.~Thananjeyan, A.~Balakrishna, J.~Ichnowski, E.~Novoseller, M.~Hwang, M.~Laskey, J.~E. Gonzalez, and K.~Goldberg, ``Untangling dense non-planar knots by learning manipulation features and recovery policies,'' in \emph{Robotics Science and Systems (RSS)}, 2021.

\bibitem{sutton2018reinforcement}
R.~S. Sutton and A.~G. Barto, \emph{Reinforcement learning: An introduction}.\hskip 1em plus 0.5em minus 0.4em\relax MIT press, 2018.

\bibitem{lillicrap2015continuous}
T.~P. Lillicrap, J.~J. Hunt, A.~Pritzel, N.~Heess, T.~Erez, Y.~Tassa, D.~Silver, and D.~Wierstra, ``Continuous control with deep reinforcement learning,'' \emph{arXiv preprint arXiv:1509.02971}, 2015.

\bibitem{haarnoja2023learning}
T.~Haarnoja, B.~Moran, G.~Lever, S.~H. Huang, D.~Tirumala, M.~Wulfmeier, J.~Humplik, S.~Tunyasuvunakool, N.~Y. Siegel, R.~Hafner, \emph{et~al.}, ``Learning agile soccer skills for a bipedal robot with deep reinforcement learning,'' \emph{arXiv preprint arXiv:2304.13653}, 2023.

\bibitem{zhou2023learning}
W.~Zhou and D.~Held, ``Learning to grasp the ungraspable with emergent extrinsic dexterity,'' in \emph{Conference on Robot Learning}.\hskip 1em plus 0.5em minus 0.4em\relax PMLR, 2023, pp. 150--160.

\bibitem{thananjeyan2021recovery}
B.~Thananjeyan, A.~Balakrishna, S.~Nair, M.~Luo, K.~Srinivasan, M.~Hwang, J.~E. Gonzalez, J.~Ibarz, C.~Finn, and K.~Goldberg, ``Recovery rl: Safe reinforcement learning with learned recovery zones,'' \emph{IEEE Robotics and Automation Letters}, vol.~6, no.~3, pp. 4915--4922, 2021.

\bibitem{wilcox2022ls3}
A.~Wilcox, A.~Balakrishna, B.~Thananjeyan, J.~E. Gonzalez, and K.~Goldberg, ``Ls3: Latent space safe sets for long-horizon visuomotor control of sparse reward iterative tasks,'' in \emph{Conference on Robot Learning}.\hskip 1em plus 0.5em minus 0.4em\relax PMLR, 2022, pp. 959--969.

\bibitem{vats2023efficient}
S.~Vats, M.~Likhachev, and O.~Kroemer, ``Efficient recovery learning using model predictive meta-reasoning,'' in \emph{2023 IEEE International Conference on Robotics and Automation (ICRA)}.\hskip 1em plus 0.5em minus 0.4em\relax IEEE, 2023, pp. 7258--7264.

\bibitem{reichlin2022back}
A.~Reichlin, G.~L. Marchetti, H.~Yin, A.~Ghadirzadeh, and D.~Kragic, ``Back to the manifold: Recovering from out-of-distribution states,'' in \emph{2022 IEEE/RSJ International Conference on Intelligent Robots and Systems (IROS)}.\hskip 1em plus 0.5em minus 0.4em\relax IEEE, 2022, pp. 8660--8666.

\bibitem{chernova2014robot}
S.~Chernova and A.~L. Thomaz, \emph{Robot learning from human teachers}.\hskip 1em plus 0.5em minus 0.4em\relax Morgan \& Claypool Publishers, 2014.

\bibitem{chi2023diffusion}
C.~Chi, S.~Feng, Y.~Du, Z.~Xu, E.~Cousineau, B.~Burchfiel, and S.~Song, ``Diffusion policy: Visuomotor policy learning via action diffusion,'' in \emph{Robotics: Science and Systems XIX, Daegu, Republic of Korea, July 10-14, 2023}, 2023.

\bibitem{ross2011reduction}
S.~Ross, G.~Gordon, and D.~Bagnell, ``A reduction of imitation learning and structured prediction to no-regret online learning,'' in \emph{Proceedings of the fourteenth international conference on artificial intelligence and statistics}.\hskip 1em plus 0.5em minus 0.4em\relax JMLR Workshop and Conference Proceedings, 2011, pp. 627--635.

\bibitem{wang2022tli}
Y.~Wang, N.~Figueroa, S.~Li, A.~Shah, and J.~Shah, ``Temporal logic imitation: Learning plan-satisficing motion policies from demonstrations,'' in \emph{Conference on Robot Learning, CoRL 2022, 14-18 December 2022, Auckland, New Zealand}, 2022.

\bibitem{pateria2021hierarchical}
S.~Pateria, B.~Subagdja, A.-h. Tan, and C.~Quek, ``Hierarchical reinforcement learning: A comprehensive survey,'' \emph{ACM Computing Surveys (CSUR)}, vol.~54, no.~5, pp. 1--35, 2021.

\bibitem{konidaris2009skill}
G.~Konidaris and A.~Barto, ``Skill discovery in continuous reinforcement learning domains using skill chaining,'' \emph{Advances in neural information processing systems}, vol.~22, pp. 1015--1023, 2009.

\bibitem{bagaria2021robustly}
A.~Bagaria, J.~Senthil, M.~Slivinski, and G.~Konidaris, ``Robustly learning composable options in deep reinforcement learning,'' in \emph{Proceedings of the 30th International Joint Conference on Artificial Intelligence}, 2021.

\bibitem{sharma2020learning}
M.~Sharma, J.~Liang, J.~Zhao, A.~LaGrassa, and O.~Kroemer, ``Learning to compose hierarchical object-centric controllers for robotic manipulation,'' in \emph{4th Conference on Robot Learning, CoRL 2020, 16-18 November 2020, Virtual Event / Cambridge, MA, {USA}}, 2020.

\bibitem{nasiriany2022augmenting}
S.~Nasiriany, H.~Liu, and Y.~Zhu, ``Augmenting reinforcement learning with behavior primitives for diverse manipulation tasks,'' in \emph{2022 International Conference on Robotics and Automation (ICRA)}.\hskip 1em plus 0.5em minus 0.4em\relax IEEE, 2022, pp. 7477--7484.

\bibitem{rosen2022role}
E.~Rosen, B.~M. Abbatematteo, S.~Thompson, T.~Akbulut, and G.~Konidaris, ``On the role of structure in manipulation skill learning,'' in \emph{CoRL 2022 Workshop on Learning, Perception, and Abstraction for Long-Horizon Planning}, 2022.

\bibitem{pomdp2010ong}
\BIBentryALTinterwordspacing
S.~C.~W. Ong, S.~W. Png, D.~Hsu, and W.~S. Lee, ``{POMDPs for Robotic Tasks with Mixed Observability},'' in \emph{{Robotics: Science and Systems V}}.\hskip 1em plus 0.5em minus 0.4em\relax The MIT Press, 07 2010. [Online]. Available: \url{https://doi.org/10.7551/mitpress/8727.003.0027}
\BIBentrySTDinterwordspacing

\bibitem{sutton1999between}
R.~S. Sutton, D.~Precup, and S.~Singh, ``Between mdps and semi-mdps: A framework for temporal abstraction in reinforcement learning,'' \emph{Artificial intelligence}, vol. 112, no. 1-2, pp. 181--211, 1999.

\bibitem{pastor2012towards}
P.~Pastor, M.~Kalakrishnan, L.~Righetti, and S.~Schaal, ``Towards associative skill memories,'' in \emph{2012 12th IEEE-RAS International Conference on Humanoid Robots (Humanoids 2012)}.\hskip 1em plus 0.5em minus 0.4em\relax IEEE, 2012, pp. 309--315.

\bibitem{chen2016xgboost}
T.~Chen and C.~Guestrin, ``Xgboost: A scalable tree boosting system,'' in \emph{Proceedings of the 22nd acm sigkdd international conference on knowledge discovery and data mining}, 2016, pp. 785--794.

\bibitem{schulman2017proximal}
J.~Schulman, F.~Wolski, P.~Dhariwal, A.~Radford, and O.~Klimov, ``Proximal policy optimization algorithms,'' \emph{arXiv preprint arXiv:1707.06347}, 2017.

\bibitem{raffin2019stable}
A.~Raffin, A.~Hill, M.~Ernestus, A.~Gleave, A.~Kanervisto, and N.~Dormann, ``Stable baselines3,'' 2019.

\bibitem{robosuite2020}
Y.~Zhu, J.~Wong, A.~Mandlekar, and R.~Mart\'{i}n-Mart\'{i}n, ``robosuite: A modular simulation framework and benchmark for robot learning,'' in \emph{arXiv preprint arXiv:2009.12293}, 2020.

\end{thebibliography}

\end{document}